\documentclass[11pt, oneside]{article}   	% use "amsart" instead of "article" for AMSLaTeX format
\usepackage[margin=0.69in]{geometry}                		% See geometry.pdf to learn the layout options. There are lots.
\usepackage{graphicx}				% Use pdf, png, jpg, or eps§ with pdflatex; use eps in DVI mode
\usepackage{amssymb}
\usepackage{amsmath}
\usepackage{hyperref}
\usepackage[table,xcdraw]{xcolor}

\renewcommand{\vec}[1]{\mathbf{#1}}
 \definecolor{darkblue}{rgb}{0, 0, 4}
  \hypersetup{colorlinks=true,citecolor=darkblue, linkcolor=darkblue, urlcolor=darkblue}
  \newcommand\tab[1][1.25cm]{\hspace*{#1}}
\title{Structured Prediction in NLP - A survey}

\author{
    Chauhan Dev \qquad   
    Naman Biyani \qquad
    Nirmal P. Suthar  \\
    Prashant Kumar \qquad
    Priyanshu Agarwal \qquad   \\
   {\tt \{devgiri, namanb, nirmalps, praskr, priyanag\}@iitk.ac.in}\\
{Indian Institute of Technology Kanpur (IIT Kanpur)}
}

\date{}							% Activate to display a given date or no date
\vspace{-0.7cm}
\begin{document}
\maketitle
\vspace{-0.8cm}
\abstract{Over the last several years, the field of Structured prediction in NLP has had seen huge advancements with sophisticated probabilistic graphical models, energy-based networks, and its combination with deep learning-based approaches. This survey provides a brief of major techniques in structured prediction and its applications in the NLP domains like parsing, sequence labeling, text generation, and sequence to sequence tasks. We also deep-dived into energy-based and attention-based techniques in structured prediction, identified some relevant open issues and gaps in the current state-of-the-art research, and have come up with some detailed ideas for future research in these fields. }
% \NB{Abstract ke liye koi bhi paper ka line mile achi to add kardena}
\vspace{-0.3cm}
\section{Introduction}
\vspace{-0.2cm}
% \NB{Intro ke liye koi bhi paper ka line mile achi to add kardena}
% Some guidelines: 

% \noindent In this section, you introduce the topic and explain in detail as if it could be understood by anyone who doesn't know about it. Give general overview of the field, motivation, applications, etc. Towards the end of introduction you also describe briefly the structure of the report, what will you cover, what is there in what section, etc.

In structured prediction, we seek to predict structured outputs, which are essentially anything other than a scalar or categorical quantity. Many prediction problems in natural language processing, such as tagging, parsing, coreference resolution, and machine translation, require the prediction of outputs that are structured. For example, predicting a sequence of part-of-speech labels, a parse tree, a partition of a document’s noun phrases, or a well-formed sentence, respectively. Intuitively, these sorts of predictions differ from those made in multi-class classification, in that the objects being predicted have an internal structure (unlike the discrete labels multi-class classification problems). This structured output is often found to be a useful intermediate representation for other applications. For example, downstream reasoning about the content of a news article may require predicting relational data about the individuals and events discussed in the text. The number of possible outputs for a particular input x also tends to be exponential in the size of x, which makes the problem challenging. Another major challenge in structured prediction is that often the inference becomes intractable. 
% In NLP, these may be a sequence of labels or graphs for example. 

% Many prediction problems in natural language processing, such as tagging, parsing, coreference resolution, and machine translation, require the prediction of outputs that are structured. 
% In these cases, for instance, we are interested in predicting a sequence of part-of-speech labels,
% a parse tree, a partition of a document’s noun phrases, or a well-formed sentence, respectively. Intuitively, these sorts of predictions differ from those made in multi-class classification, in
% that the objects being predicted have internal structure (unlike the discrete labels we predict
% in multi-class classification problems).  Consequently, the number of candidate outputs for a
% particular input x also tends to be exponential in the size of x which makes the problem of structured prediction challenging. Another major challenge in structured prediction is that often the ineference becomes intractable
% Also, the structured output may be the output of a content generation system used as the interface between a computer and a user. For example, when a dialogue system responds to a user query, it may produce its response as a sentence containing multiple words, and this sentence may be further converted into an audio signal for a simulated person speaking the sentence.

% TODO:
% \begin{itemize}
%     \item 
% \end{itemize}
\vspace{-0.3cm}
\section{HMMs, MEMMs, CRFs and Structured Perceptron}
\vspace{-0.2cm}
\textbf{Hidden Markov Models} \cite{hmm}
is a probabilistic graphical model which uses first order markov chain assumption and dependence of observed data only on emission states. HMM allows us to talk about both observed events and hidden events in our probabilistic model. The probability of a sequence(length=$T$) is modeled as follows: 
$$
p(y|x,\theta) \propto  p(x,y,\theta) = p(y_1)\prod_{i=2}^{T}p(y_{i}|y_{i-1})\prod_{i=1}^{T}p(x_{i}|y_{i})
$$
The forward algorithm allows to efficiently calculate the MLE or MAP estimate and 
Viterbi-style dynamic programming algorithm \cite{viterbi} can be used for inference. A clear drawback is that transition probabilities and emission probabilities do not encode any information other than the previous tag and also HMMs suffer from exposure bias problem and hence are often found to be quite inefficient.

\textbf{Maximum Entropy Markov Models} (MEMMs) \cite{McCallum00maximumentropy}
define a probabilistic model of sequences $y \in Y = V^T$ given
an input sequence $x$, also of length $T$. The probability of a sequence is modeled as the product of locally normalized per-label distributions, as follows:
\vspace{-0.2cm}
$$
p(y|x) = \prod_{t=1}^Tp(y_t|x_t,y_{t-1}) = \prod_{t=1}^T \frac{exp(w_t^T \phi_t(x_t,y_t,y_{t-1}))}{\sum_{v \in V}exp(w_t^T \phi_t(x_t,v,y_{t-1}))}
$$

MMEMs assume a first-order Markov assumption which means it assumes that the label $y_{t}$ given label $y_{t-1}$ is independent of all the previous labels.
% \PK{changed markov definition}
This independence assumption allows for the arg max sequence
to be calculated exactly with a Viterbi-style dynamic programming algorithm, similar to the inference in Hidden Markov Models. 
Also, this independence assumption gives features which only depend on $x_t$,
$y_t$, and $y_{t-1}$ for each $t$. 
% MEMMs based approached have been founA Maximum Entropy Approach to Natural Language \cite{berger-etal-1996-maximum} .\NB{Add some lines to cite this}
As is evident from probability of sequence equation, the probability of a sequence $y$ for an input $x$ is the product of locally-normalized per-label probabilities. This leads to a problem known as label bias. 

\textbf{Condition Random Fields}
\cite{crf} which uses global normalization and specifies the conditional distribution of a random variable $Y$ given a random variable $X$ with respect to the factorization implied by an undirected graphical model over $X$, $Y$. Distribution in a CRF following first-order markov property can be written as follows:
\vspace{-0.2cm}
$$
p(y|x) = \frac{\prod_{t=1}^T exp(w_t^T \phi_t(x_t,y_t,y_{t-1}))}{\sum_{y' \in Y}\prod_{t=1}^T exp(w_t^T \phi_t(x_t,y_t,y_{t-1}))}
$$
% \PK{removed partition function line}
% The denominator is also known as the partition function.
Unlike MEMMs, the distribution $p(y | x)$ is not the product of
locally normalized distributions. It is rather globally normalized, using the partition function, and so this model does not suffer from label bias. 
% \PK{check woding here $\to$}
Given the Markov assumptions implicit in the model above, we may again find the arg max
sequence using a Viterbi-style algorithm.
 \cite{sutton2010introduction} has explained that computing the arg max, marginals or the partition function is usually not algorithmically efficient.
 
 \textbf{Structured Perceptron} \cite{Dirscherl_1998} and \textbf{structured SVM} \cite{Wu_2008} based approaches only require efficiently compute the arg max, unlike CRFs which require computing many other marginals and partition function. These approaches are non-probabilistic, and they utilize a simple form for the score:
 \vspace{-0.25cm}
 $$
p(y|x) = {\sum_{t=1}^Tw_t^T \phi_t(xy)}
$$
Structured Perceptron and structured SVM are trained through gradient based approaches using margin and slack rescaling losses. These approaches do not suffer from exposure
bias and label bias problem. % as they do not assume a gold history during training and are not locally normalized.

% Finally, we note that the Structured Perceptron and SSVMs suffer from neither exposure
% bias nor label bias, since they do not assume a gold history during training, and they do not
% calculate locally normalized scores. Furthermore, the ∆ terms theoretically allow a loss that
% does not decompose over parts of yˆ, though, as noted above, a non-decomposable loss might
% make calculating the arg max intractable.
% \NS{can add structured SVM and perceptron}
\vspace{-0.3cm}
\section{Structured Prediction Energy Networks}
\vspace{-0.2cm}
% \PK{Draft of SPEN paper}

% * SPEN
% ! Features
% SPEN was the first model to use deep architecture to model energy of labels.
% Most of the prev approaches assumed a particular graphical models for energy $E_x(.)$ which results in  an excessively strict inductive bias. 
% Inference is done by approximate energy minimization using algorithms like gradient descent. SPENs were found to be very well suited to multi-label classification problems.

% ! arch
Energy based approaches rely on the principle that there is an energy function which assign values to each output $\vec{y}$ given an input $\vec{x}$ such that the most compatible $\vec{y}$ has the minimum energy. Additive nature of energy makes it straightforward to incorporate domain knowledge and hard constraints in the model. Probability based approaches are a subset where probabilities can be thought as normalised energy. Most of the previous approaches assumed a particular graphical models for energy $E_\vec{x}(.)$ which results in an excessively strict inductive bias. 

\textbf{Structured prediction energy networks (SPENs)} was introduced in \cite{spen1} as the first framework for structured prediction in which energy is defined using neural networks over possible outputs and gradient based approaches for energy minimization is used for inference. SPENs were found to be very well suited to multi-label classification problems in the NLP domain.
The neural network consist of two arbitrary neural network $F(\vec{x})$ (feature network) and $E(F(\vec{x}), \vec{\bar{y}})$ (energy network). Energy network is sum of two terms, $E_\vec{x}(\vec{\bar{y}}) = E_{\vec{x}}^{local}(\vec{\bar{y}}) + E_\vec{x}^{global}(\vec{\bar{y}})$ where
\vspace{-0.25cm}
$$
  E_{\vec{x}}^{local}(\vec{\bar{y}}) = \sum_{i=1}^{L}\bar{y}_i \vec{b_i}F(\vec{x}) \tab
  E_{\vec{x}}^{global}(\vec{\bar{y}}) = \vec{c_2}^T g(\vec{C_1}\vec{\bar{y}})
$$
% \vspace{-0.1cm}
% !  training
The SSVM (structured support vector machine) technique is used for training and the loss is :-
\vspace{-0.25cm}
\begin{equation}
  \sum_{\{\vec{x_i}, \vec{y_i}\}} \max_\vec{y} [ \Delta(\vec{y_i}, \vec{y}) - E_{\vec{x_i}}(\vec{y}) + E_{\vec{x_i}}(\vec{y_i}) ]_{+}
  \label{eq:spen-ssvm}
\end{equation}
% \vspace{-0.1cm}
Where $ \Delta(\vec{y_i}, \vec{y})$ is a error function between prediction and ground truth label and $[.]_{+} = \max(0,.)$. Sub-gradient computation requires \textit{loss-augmented inference}: $\label{eq:spen-lai} \vec{y_p} = \arg \min_\vec{y} (-\Delta(\vec{y_i},\vec{y}) + E_\vec{x_i}(\vec{y}))$ for every training example.
% ! comparisons and shortcomings, future work
SPENs are more expressive and scalable while being less vulnerable to overfitting and performing competitively to previous energy-based approaches.
Like most energy networks, the inference involves optimization of $E_\vec{x}(\vec{y})$ w.r.t $\vec{y}$ which limits the practical use of model to an extent. Also, non-convexity of $E_\vec{x}(.)$ or relaxation used to get $\vec{\bar{y}}$ from $\vec{y}$ can affect the model performance and convergence. Some of these can be tackled using techniques like \textit{inference network} \cite{tu2018learning} and \textit{Input Convex Neural Networks} \cite{pmlr-v70-amos17b}.

\textbf{Deep value networks(DVN)}, introduced in \cite{gygli2017deep}, are similar to SPENs but are much easier to train and works even when $\Delta(\vec{y},\vec{y^*})$ is non-continuous and require no convexity constraints.
% Inference strategy similar to that of the original SPEN architecture \cite{spen1}. 
The key idea is that the loss function in every structured prediction problem can be used to define a \textit{quantitative value} for a configuration as $v(\vec{x}, \vec{y}) \approx - \Delta (\vec{y}, \vec{y^*})$. The deep network will model $v(.,.)$ and hence learn to mimic the loss function. Instead of using max-margin surrogate objective (SSVM) \cite{spen1} or other complex approaches, this method just do a simple regression during training and has better performance than \cite{spen1}.  Training data ($(\vec{x}, \vec{y}, v(\vec{x}, \vec{y}))$ tuples) can be generated via random sampling, generating adversarial tuple, or samples generated during train-time inference. Therefore it needs lesser number of $(\vec{x}, \vec{y^*})$ ground truth pairs for training. Gradient based inference (deep dreaming) is found to iteratively refine prediction.

SPEN objective (Eq. (\ref{eq:spen-ssvm})) optimisation requires loss augmented inference which may fail to discover margin violations if exact energy minimisation is intractable. This can be tackled using \textbf{End-to-End learning for SPENs}, introduced in \cite{belanger2017endtoend} where the energy function is discriminatively trained by back-propagating
through gradient-based prediction. These are more accurate and allows non-convex energy. The idea is not to do exact energy minimization and instead assume a particular approximate minimisation algorithm and optimize performance of this algorithm.
Suppose we chose $T$ step gradient descent as our algorithm for minimization where $T$ is some hyper-parameter. Then, the model's output is $\vec{y_T}$ given by,
\vspace{-0.3cm}
\begin{equation}
    \vec{y_T} = \vec{y_0} - \sum_{t = 1}^{T} \eta_t \frac{d}{d \vec{y}}E_\vec{x} (\vec{y_t})
    \label{eq:e2eSPEN}
\end{equation}
The energy function is defined identically to the original SPEN paper. However, energy funciton must be twice sub-differentiable and it is not straightforward to implement this model using standard learning libraries. The model loss can be defined as, $L = \frac{1}{T}\sum_{t = 1}^{T} w_t \Delta(\vec{y_t}, \vec{y^*})$ with $w_t = \frac{1}{T-t+1}$.
Unlike in DVNs, here explicit energy function provides interpretability. Another benefit is that the model outputs inference procedure along with the energy function. The inference procedure has fixed $T$ meaning that the inference time complexity is independent of $\vec{x}$.
Although it is worth noting that the learned energy may be such that the energy minima doesn't actually represents the best output as our model is learning just to optimise the inference procedure which may not be converging to minima.

% * Training spen with indirect suprevision
The approaches used till now for training SPENs were supervised and required labeled data which are expensive to collect in many problems, so we are now exploring the space of semi-supervised learning where we use domain knowledge in form of reward functions and ranking based supervision.
The first approach introduced in \cite{rooshenas-etal-2018-training} is to rank every pair of consecutive candidates obtained from training time gradient-descent inference on $\vec{y}$; and update energy network parameters by optimising an objective function if the energy values are not consistent with domain knowledge based ranking. 
In the above approach we collect training pairs from gradient descent based inference on energy functions, but it has several problems such as not able to get relevant pairwise rank violations, not reaching to a region of high reward or if we have plateaus in our energy functions. To overcome these issues, \cite{rooshenas2019searchguided} introduced a new approach where for each example we start with a gradient descent inference output, and then use a randomized search procedure for iteratively selecting a random output variable, if the reward increases more than the margin we return the new sample or else we begin sampling again.

\textbf{Graph Structured Prediction
Energy Networks (GSPEN)} \cite{graph_spen} was introduced as a combination of classical structured prediction models and SPENs to score an
entire set of predictions jointly which allows to both model explicit local and implicit higher-order correlations while maintaining tractability of inference. Inference in GSPENs is a maximization of a generally non-concave function with respect to structural constraints which is an approach inspired by \cite{NEURIPS2018_a2d10d35}. GSPENs use Frank-Wolfe Inference as its objectives
% \PK{yaha ',' hoga kya?} 
are non-linear
while the constraints are linear and use Structured Entropic Mirror Descent for optimization, an approach similar to SPEN optimization with some additional entropy-based constraints.
GSPENs are found to outperform SPENs and other baselines on multilabel classification and named entity recognition.
% kaunsa cite karna h search guided loighlty subtap?
% ervised ,,,,,,,ek aur paper hai usko bibtex mein daal de.....training spen with indirect supervision {Rooshenas 2017}..isko add kr dena are rooshena wala upr dalana\usepackage{} the first approach

%https://meet.google.com/vbd-havd-mja

% Unocomment after reading paper
% Classical probabilistic graph based approaches for structured prediction either suffers from lack of ability to model high-order correlations among variables while being computationally tractable or cannot model explicitly model known correlations. So, Graph Structured Prediction Energy Networks\cite{graph_spen} which proposed inference techniques that could model both explicit local and implicit higher-order correlations alongside being tractable.

% Deep energy-based models for structured prediction \cite{belanger2017deep}
% Search-Guided, Lightly-supervised Training of Structured Prediction Energy Networks\cite{rooshenas2019searchguided}
% Improving Joint Training of Inference Networks and Structured Prediction Energy Networks\cite{tu2020improving}

\vspace{-0.3cm}
\section{Structured Attention networks}
\vspace{-0.2cm}
Attention networks are now a standard part of the deep learning toolkit and are found effective approach for embedding categorical inference but they often fail to model richer structural dependencies in end-to-end training. Structured Attention networks\cite{kim2017structuredattention} was proposed as a simple extension of the basic attention procedure which allow for extending attention to partial segmentations or to subtrees. They explain how they can implement a linear-
chain conditional random field and a graph-based parsing model as classes of structured attention networks, which are equivalent to neural network layers.
It is shown that provide a structural bias by modelling structural dependencies at the
final output layer, most notably in seminal work on graph-based neural architectures\cite{lample-etal-2016-neural}, \cite{collobert2011natural}.

In the structured attention model, the attention distribution $p(z|x, q)$ is defined as conditional random field (CRF) or other graph based model to specify independent structure of $z$: a vector of discrete latent variables. The CRF is parameterized with clique log-potentials $\theta_C(z_C)\in R$ where the $z_C$ indicates subset of $z$ given by clique $C$, and attention probability is $p(z|x, q;\theta) = \text{softmax}(\sum_C \theta_C(z_C))$ where $x$ is the input sequence and $q$ is the query.  In practice we use a neural CRF, where $\theta$ comes from a deep model over $x$, $q$. Here also, it is assumed that annotation function $f$ factors into clique
annotation functions $f (x, z) = \sum_C f_C(x,z_C)$ and the context is defined as $c = E_{z \sim p(z|x,q)}[f(x,z)] =  \sum_{C}E_{z\sim p(z_C | x,q)}[f_C(x,z_C)]$. In short,the attention mechanism is taking expectation of an annotation function
$f (x, z)$ with respect to a latent variable $z \sim p$,
where $p$ is parameterized as function of $x$ and $q$. Instead of linear-chain CRF, syntactic tree-based attention can also be used for modelling richer structural dependencies using similar approach. Also, unlike previous approaches which add Graphical models as the final layer, they proposes that these can be added within deep networks in place of simple attention layers.

Using the two ways mentioned for structured attention, they experiment on a variety of synthetic and real tasks: neural machine translation, question answering, and natural language inference and found that it outperformed baseline attention models. They also conclude that structured attention leads to learning interesting unsupervised hidden representations that generalize simple attention.
% Structured attention networks\cite{kim2017structuredattention}
Structural attention network for graph  \cite{graphattention} is a novel approach to learn node representations for graph modelling. It uses a graph-based attention network, to pay attention to the topology of the graph. 

Graph classification is a problem with many practical applications and an approach \cite{graphclassification} using structural attention was proposed recently. It's the first model to uses attention to learn representations for general attributed graphs.
% The use of attention is found to focus on informative parts
% of the graph and avoid graph noise. 
A major problem in graph-based classification was that graphs are mostly noisy and large and processing the entire graph can inadvertently caused noise to be introduced into the calculated feature and also computationally inefficient which increased the need for structured attention in graphs.
The approach used an RNN based model that
processes only a portion of the graph by adaptively selecting a
sequence of “informative” nodes. Furthermore, the attention-guided
walk mechanism relies only on local information from the graph
which  keeps computation costs  low since there is no need to load the
entire graph into memory. With various experiments on real-world datasets, they concluded that the proposed method is competitive
against various well-known methods in graph classification even
though our method is limited to only a portion of the graph.
% Graph Classification using Structural Attention
% ye ek achi link h for parsing survey https://arxiv.org/pdf/1807.10854.pdf

\vspace{-0.3cm}
\section{Tree LSTM}
\vspace{-0.2cm}
Tree LSTMs \cite{tai2015improved} were proposed to generalize LSTMs (chain-structured) to tree-structured network topologies and they were found to outform all LSTM baselines on predicting semantic relatedness of two sentences and
sentiment classification and hence found superior for
representing sentence meaning.  
% While the standard LSTM composes its
% hidden state from the input at the current time
% step and the hidden state of the LSTM unit in the
% previous time step, the tree-structured LSTM, 
Tree-LSTM, composes its hidden state from an input vector and the hidden states of arbitrarily many child units unlike hidden state of previous time step in normal LSTM.

Recently, Tree-LSTMs have been widely applied in structured prediction tasks like dependency and constituency parsing. Works on Recursive LSTM Tree have been proposed for feature Representation of parse trees in dependency and constituency parsing \cite{lstm_tree_parsing}, \cite{kirnap-etal-2018-tree}, \cite{kleenankandy2020enhanced}. These approaches have removed the need of handcrafted structural features and have shown better performance as well.

Recently, many works in the community have tried to improve Tree-LSTMs \cite{ahmed2019improving} proposed encoding variants of decomposable
attention inside a Tree-LSTM cell for dependency
and constituency trees. To avoid the bias problem of the root node dominance, capsule tree-LSTM \cite{wang-etal-2019-investigating} was proposed, which uses dynamic routing
algorithm to build
sentence representation by assigning
different weights to nodes according to
their contributions to prediction.
% Recursive LSTM Tree Representation for Arc-Standard Transition-BasedDependency Parsing\cite{lstm_tree_parsing}\
    % TreeLSTM with tag-aware hypernetwork for sentence representation \cite{tagaware}
\vspace{-0.3cm}
\section{Application of Structured Prediction techniques in NLP}
\vspace{-0.2cm}
\subsection{NER, POS and other Sequence Labelling tasks}
\vspace{-0.2cm}
% \NB{ NER POS :- Naman}
Linguistic sequence labeling tasks like POS tagging and named entity recognition (NER), are one of the first stages in language understanding and its importance has been well recognized. Many NLP systems like coreference resolution and syntactic parsing are becoming more sophisticated by utilizing output information of POS tagging or NER.

Linear statistical models like HMMs, MEMMs and CRFs give high performance on such sequence labelling tasks by using hand-crafted features and task-specific resources like English POS taggers benefit from carefully designed word spelling features. However, such task-specific knowledge is costly
to develop. Due to this, deep-learning based approaches which use word embeddings as features are being applied to these tasks and are giving great results. Several RNN together with its variants such as long-short term memory (LSTM) \cite{lstm} and
gated recurrent unit (GRU) \cite{gru} have been proposed to solve sequence labeling
tasks like POS tagging \cite{pos-neural} and NER \cite{chiu2016named}, \cite{hu2020harnessing}. As these models depend a lot on neural embeddings, often its performance drops rapidly and so new approaches combining linear statistical models on top of RNN-based models. Approaches like  \cite{huang2015bidirectional},\cite{nercrf} combine character and word-level representations, feed them into a BiLSTM and then use a CRF on top of it to jointly decode labels for the whole sentence and they give great accuracies in sequence labelling tasks. Hierarchical RNN based encoders along with CRFs have also been used for sequence labelling \cite{kumar2017dialogue}
\vspace{-0.2cm}
\subsection{Semantic Parsing and Semantic analysis}
\vspace{-0.2cm}
Semantic Processing involves the understanding the meaning of phrases, sentence or documents. Research have directed attention towards learning word representation methods from a large unlabeled corpus and claim to capture the meaning of words such as Word Embedding \cite{mikolov2013efficient} and GloVe \cite{pennington-etal-2014-glove}. Research have also shown that \cite{Liu2018SemanticSW} semantic structures helps in defining effective word embedding. SENSE \cite{Liu2018SemanticSW} uses hierarchically organized data that could be used in generating semantic structures to exploit the semantic relationship such as parent, brother and child semantic attributes to learn effective embedding. This explicitly uses the external semantic structured information to learn the word representations by optimizing the prediction ability between target word and context.   

Semantic Parsing is task of mapping the natural language to logical representation on which we can support various application such as machine translation, question answering, etc. Semantic role labeling, a second-order structured prediction task is one such application of semantic parsing also known as Shallow semantic parsing, in which we are concerned with labelling phrases of sentences with its corresponding semantic roles, but not encovering deeper compositional semantics of text. Approached based on contextualized embeddings, such as ELMo \cite{elmo}, BERT \cite{bert}, XLM-R \cite{xlmr} have been emerged as state-of-the-art for various structured prediction tasks such as semantic role labelling, chunking etc. Research have also showed that word representation based on the ensembling or concatenation of multiple pretrained embedding and non-contextual embedding can further improve the performance of structured prediction tasks \cite{FLAIRS2018438}.

% Bidirectional LSTM-CRF Models for Sequence Tagging\cite{huang2015bidirectional}
% End-to-end Sequence Labeling via Bi-directional LSTM-CNNs-CRF\cite{nercrf}
% Dialogue Act Sequence Labeling using Hierarchical encoder with CRF\cite{kumar2017dialogue}

% Neural Models for Sequence Chunking\cite{neuralchunking}
% Learning Structured Perceptrons for Coreference Resolution with Latent Antecedents and Non-local Features\cite{bjorkelund-kuhn-2014-correference}An Empirical Investigation of Beam-Aware Training in Supertagging\cite{negrinho2020empirical}
\vspace{-0.2cm}
\subsection{Structured Prediction in Dependency Parsing}
\vspace{-0.2cm}
% \NS{ Parsing :- Nirmal }

% % Parsing examines how different words and phrases relate to
% % each other within a sentence.  There are majorly two
% % forms of parsing:  Dependency and Semantic parsing.
% Dependency parsing
% looks at the relationships between pairs of individual words.

In recent year, there have been more focus transition-based approaches for parsing, which generates one parse tree. Arc-standard parsing starts with the an empty stack and a queue consisting of whole input sequence. At each step, a word input is consumed and corresponding stack operations are applied. At the end whole sentence is consumed and stack contain only ROOT label.
Transition based methods have been giving top accuracies for dependency parsing \cite{huang-sagae-2010-dynamic} \cite{zhang-nivre-2011-transition} \cite{goldberg-nivre-2013-training}. Neural parsing with global learning and beam search have also come out to be start-of-the-art transition-based parsers \cite{zhang-nivre-2012-analyzing}, which also leverages structured-prediction model to optimize whole sequence of stack actions \cite{zhang-nivre-2011-transition} instead of using local classifier that optimizes each action. There have been efforts in proposing structured prediction model for probabilistic  dependency parsing \cite{zhou-etal-2015-neural}, which maximizes the likelihood of arc-standard based action sequence instead of individual actions. 
\vspace{-0.3cm}
$$ p(y_i|x,\theta) = \frac{e^{f(x, \theta)_i}}{\sum_{y_j \in GEN(x)} e^{f(x, \theta)_j}}  \quad \text{where  }  f(x,\theta)_i = \sum_{a_k \in y_i} o(x, y_i, k, a_k) $$

The probability of the action sequence $y_i$ is given by above equation. Here $o(x, y_i, k, a_k)$ denotes the neural network score for the action $a_k$ given $x, y_i$. The probability captures the likelihood of all action sequence instead of individual actions. Results showed that this model outperformed greedy dependency parsers, and also able to have comparable results with small parameter size.

% \NS{[AGENTShaw Publishing] offered [RECEPIENTMr. Smith] [THEMEa reimbursement] [TIMElast March]. 
% Can paste an image of example}

% Done
% Improving Semantic Parsing for Task Oriented Dialog\cite{einolghozati2019improving}

% Neural Semantic Parsing with Type Constraints for Semi-Structured Tables\cite{krishnamurthy-etal-2017-neural}

% A Neural Probabilistic Structured-Prediction Model for Transition-Based Dependency Parsing\cite{zhou-etal-2015-neural}

% Bidirectional LSTM-CRF Models for Sequence Tagging
% Zhiheng Huang, Wei Xu, Kai Yu

% TODO

% Semantic Parsing for Task Oriented Dialog using Hierarchical Representations\cite{gupta-etal-2018-semantic-parsing}
% Don’t Parse, Generate! A Sequence to Sequence Architecture for Task-Oriented Semantic Parsing\cite{Rongali_2020}
% Incremental Recurrent Neural Network Dependency Parser with Search-based Discriminative Training\cite{yazdani-henderson-2015-incremental}

% A Neural Probabilistic Structured-Prediction Model for Transition-Based Dependency Parsing\cite{zhou-etal-2015-neural}
% Sequence-to-Sequence Learning as Beam-Search Optimization \cite{beamopt}
% Transition-Based Dependency Parsing with Stack Long Short-Term Memory\cite{dyer-etal-2015-transition}

% \newpage
\vspace{-0.2cm}
\subsection{Text generation}
\vspace{-0.2cm}
The importance of global normalization for text generation was emphasized by \cite{beamopt} which has led to introduction of statistical models like CRFs combined with neural network architectures in text generation domain.
Sequential autoregressive models which predict make predictions based on past predictions are the predominant architecture for
text generation in a maximum likelihood setup but MLE based training leads to teacher forcing and exposure bias. Feedback loops have been used to overcome this issue where one provides the last
predicted token as a feature to the computation
of the next state. % but this was adopted by non-autoregressive models as well. {ye sahi likha hai kya? ki non-autoregressive model me feedback loop adopt kiya hai?}
Recently, \cite{schmidt-etal-2019-autoregressive} proposed modeling temporal correlations as
part of the observation model by a neural CRF observation model that leverages word-
embeddings to explain local word correlations in a global sequence score. The resultant autoregressive model kept hidden state evolution less affected by observation noise
while generating coherent word sequences.

% Fully-autoregressive models perform beam-searching to decode the text while non-autoregressive models does parallel decoding. 
Fully-autoregressive models are state-of-the-art in text generation but beam-searching can not be parallelized, to speedup decoding using parallel generation models assumes independence of output words which leads to problems like repetitions. Markov Transformers \cite{deng2020cascaded} proposed a slightly different architecture for modern autoregressive models such that they can parameterize hierarchical CRFs, on which we use refinement based decoding procedures which are parallelizable. Such models are partially-autoregressive, and has trade off between speedup and autoregressive nature of the model which requires finding a best configuration with speedup and performance. With distillation and good configuration this model performs comparable to other parallel decoding models, has good candidate search, and less problem of repetition.

Modeling language as a sequence of discrete tokens is a widely used approach, which is only constraint to sequence dynamics. Recently, syntax driven approach \cite{casas2020syntaxdriven} is proposed, where token generation is driven by dependency parse tree with iterative generation procedure.
% Dependency tokens are also added with word tokens, where each dependency token is a non-terminal token which is expanded and iteratively fed to the model again to generate the text.
Iterative Expansion Language Models, given a level of dependency parse tree produces tokens for next level of dependency parse tree with expansion placeholders (non-terminal tokens which can be expanded to multiple tokens for next iteration), inference is done by starting from fix root token as first level of dependency parse tree and iteratively generating all levels till terminals are generated, requiring only few iterations and beam-search can be used across iterations. This iterative non-autoregressive model generates text with diversity quality between LSTM and transformers.
\vspace{-0.2cm}
% \NB{ Dev-Pls iska kal tak try karna jitna likh sako }

% Autoregressive Text Generation Beyond Feedback Loops\cite{schmidt-etal-2019-autoregressive} 
% Cascaded Text Generation with Markov Transformers\cite{deng2020cascaded}
% Syntax-driven Iterative Expansion Language Models for Controllable Text Generation\cite{casas2020syntaxdriven}

% \subsection{Question Answering and Machine Translation}
% % \NB{Prashant,Priyanshu, Dev-Pls iska kal tak try karna jitna likh sako }
% Attention Is All You Need\cite{vaswani2017attention}

% , Energy-Based Reranking: Improving Neural Machine Translation Using Energy-Based Models \cite{bhattacharyya2021energyb  ased}

% \NB{To be completed in approx 5 pa`ges yaha tak}
% \newpage
\vspace{-0.3cm}
\section{Future Directions}
\vspace{-0.3cm}
% GLOM\cite{hinton2021represent}
\begin{enumerate}
\itemsep-0.25em
    \item Structured Attention networks \cite{kim2017structuredattention} proposed an approach to add graphical models anywhere within deep networks in place of simple attention layers. Also, we read about SPENs \cite{spen1} used as output layers for structured prediction in energy-based models and it may be interesting to explore how to use SPEN layer as as internal neural network layers in the form of attention-based modules (replace CRF based attention module with SPEN based module), using approximate inference methods and end-to-end training mechanism mentioned in \cite{kim2017structuredattention}, \cite{belanger2017endtoend}, and \cite{pmlr-v70-amos17b}. We may try finding alternatives to ICNN \cite{pmlr-v70-amos17b} approach which is less restrictive for enforcing convexity or employ a regularization method that enforces convexity as a soft constraint. Also, GLOM \cite{hinton2021represent} mentioned about consensus based attention mechanism which can be modelled using linear-chain CRF with pairwise dependencies.
    % \PK{...}
    \item Much work has not been done in the field of seq-to-seq models and  Natural language inference using structured prediction based techniques and recently it has been found that transformers \cite{vaswani2017attention} give great accuracy on these tasks but it was found in TaxiNLI \cite{joshi2020taxinli} that many taxonomic categories were found to be difficult to be predicted accurately by transformers in Natural Language inference and question answering. We want to experiment by adding structured prediction based methods on top of the output layer in such seq-to-seq tasks and see which are more robust to all taxonomic categories. Also, we would like to experiment how transformer would work if we replace its normal soft attention layer by structured attention layer.
    \item Recent structured prediction approaches that we surveyed like SPENs, DVNs and TreeLSTMs are being applied to a very small set of tasks which are simpler than majority of structured prediction tasks. We would like to extend these ideas to other language understanding tasks where we have to predict a structured output and see if these techniques improve the baselines. There may be some problems in making the model tractable but the end-to-end training mechanisms of these individual methods or application of certain restrictions suggested in ICNNs \cite{pmlr-v70-amos17b} would make the model training work properly.
    \item Many natural language understanding datasets contain annotation artifacts or biases and we would like to explore how various NLP approaches are robust to  dataset biases. We believe structured prediction techniques would result in more robust models. We would further like to work on making the models even more robust, come up with new training methods to combat bias, and extending the ideas to other language understanding tasks. 
\end{enumerate}
\vspace{-0.5cm}
\section{Conclusion}
\vspace{-0.2cm}
This survey has explored a wide variety of methods of Structured Prediction and its applications in NLP. Our goal has been to study the general usefulness of statistical, graphical, energy-based and attention-based methods to model the structural dependencies in structured prediction tasks with more focus towards such tasks in NLP domain like POS tagging, Named Entity Recognition, Semantic Role labelling and analysis, Dependency and Semantic Parsing and Text generation.
    
Linear statistical and graphical methods like CRFs are effective but output dependencies are modeled by hand, and has significant inductive bias SPENs are its further recent work is a great contribution in the step towards incorporating deep networks and energy-based approaches in structured prediction resulting in more expressive models, and have huge potential for future work in its scalability and application to more complex tasks. Structured Attention is a great approach of adding statistical models in between neural network layers and should be explored with other structured prediction based techniques as well. TreeLSTMs are found to give good accuracy on Dependency and semantic parsing and they should be used in other NLP domains as well like seq-to-seq models using hierarchical encoders and decoders made of TreeLSTMs . Structured prediction methods may have great potential when combined with Transformer \cite{vaswani2017attention} which are the current SOTA in most of the NLP tasks.
% may give +CRF/MEMM/HMM........
% There should be a conclusion section. The name is obvious, you need to conclude your report. 
% \vspace{-0.3cm}
% \section{Individual Contributions}
% \vspace{-0.3cm}
% % This should be in tabular format. Also in this briefly mention about what each member would be contributing in the mini-research project. 
% \begin{center}

% \begin{table}[h!]
% \centering
% \begin{tabular}{||c | c||}  
%  \hline
%  \textbf{Name} & \textbf{Techniques/Domains Studied} \\ 
%  \hline
%  \hline
%  Chauhan Dev &  Structured Attn, Text Generation\\ 
%  \hline
%  Naman Biyani &  TreeLSTM, Sequence and Semantic Labelling \\ 
%  \hline
%  Nirmal Suthar & Sequence and Semantic Labelling, Parsing\\
%  \hline
%  Prashant Kumar & Structured Attn, Energy networks  \\ 
%  \hline
%  Priyanshu Agarwal & Structured Attn, Energy networks \\ 
%   \hline
%  All & HMM, MEMM, CRF, Structured Perceptron and SVM \\ 
%  \hline
%  \hline
% \end{tabular}
% % \caption{Individual Contributions}
% \label{table:1}
% \end{table}
% \vspace{-0.52cm}
% \end{center}
% \vspace{-0.92cm}
\newpage
\bibliography{references} 
\bibliographystyle{ieeetr}

\end{document}